A dissertation submitted in partial fulfilment of the requirements for the University of Greenwich Master's Degree in Software Engineering

# Vehicle Speed Detecting App

**Name:**                                  Itoro Michael Ikon

**Student ID:**                          XXXXXXXX

**Programme of Study:**          MSc Software Engineering

**Supervisor:**                         XXXXXXXX

**STUDENT PLAGIARISM DECLARATION**

I hereby declare that the work submitted for assessment is original and my own work, except where acknowledged in the submission.

Signed:                                                                      Date:

Word Count: 12615

## Abstract

This report presents the measurement of vehicular speed using a smartphone camera. The speed measurement is accomplished by detecting the position of the vehicle on a camera frame using OpenCV's library LBP cascade classifier, the displacement of the detected vehicle with time is used to compute the speed. Conversion coefficient is determined to map the pixel displacement to actual vehicle distance. The speeds measured are proportional to the ground truth speeds.

## Acknowledgements

The completion of this report has been possible only because of the support received from numerous sources which deserve special mention. I would like to begin by thanking God for the grace to excel through the Master's program.

A big thank you to every member of my family: XXXXXXXX are greatly appreciated for financially sponsoring the MSc program. XXXXXXXX are acknowledged for being very supportive during the course of this Msc program. XXXXXXXX deserves an inexpressible gratitude for his benevolence and encouragement.

I thank XXXXXXXX and XXXXXXXX my supervisor and second maker respectively for their useful suggestions that have been essential to the completion of this project. XXXXXXXX the Program Leader is also acknowledged for her high sense of duty and support to all Master's students assigned to her.

My colleagues in the Master's program and in my Department have been good friends and are appreciated.

## Table of Contents

## List of Tables

# List of Figures

# 1 Introduction

The report will cover the development of the vehicle speed detecting app (referred to as Speed app). In this section, the context for the choice of the project will be covered, there will be a discussion on the objectives of the project, the methodology used in the development of the project will be considered and there will be a description of the different sections of this report.

The project has been developed to serve as a mobile speed camera running on an Android powered smartphone. A speed camera is usually positioned at the side of a road to detect the over speeding of vehicles and capture the picture of such vehicles (Dictionary.cambridge.org, 2016). The project has been synthesised for the scenario where a local resident wishes to garner evidence of over speeding in a neighbourhood so that it could be used to apply to local traffic authorities for the construction of a speed hump (Anthony, 2016; Department for Transport, 2013). The Speed app is also designed to be used by Police constabularies as a backup camera (Windall, 2016). Some motorist do over speed in between established speed camera points, hence mobile speed cameras based on laser technology are mounted on vans and deployed by law enforcement officers to apprehend such law breakers (Racfoundation.org, 2016). The Speed app serves as a low cost backup mobile speed camera to the digital ones currently in use which have an average cost of £20,000 (Racfoundation.org, 2016).

To achieve the quest for a low cost mobile speed camera, the vehicle speed detecting app project has the following as its objectives:

a. Research on existing object recognition techniques and their APIs.
b. Determine the most effective object recognition API to incorporate.
c. Design and implement an Android app utilising the chosen API to detect vehicle speed, the app will also capture time and location of capture.
d. Back up app data to a web service hosted on the University's IIS server
e. Test the developed system to determine effectiveness and complete the app specification
f. Suggests other ways the developed system could be further improved.

Objective a and b will be covered in the Literature Review section, the object detecting API incorporated into the app is the Open Source Computer Vision Library (OpenCV) for Android version 2.4.11 (Opencv.org, 2016). The project is utilising its implementation of the Local Binary Pattern (LBP) classifier algorithm (Liao et al., 2007). The design of the Speed app will be detailed in the section on App Analysis and Design, the actual implementation of the app in code will be presented in the Implementation section. The backing up of the data generated by the app to a web service is a part of the overall project that will be considered in the aforementioned sections. The results of testing the app will be relayed in the section on System Testing, and finally, recommendations will be made in the Conclusions and Recommendations section on ways through which the app can be improved. In the Appendices section the design artefacts used to realise the system are placed.

With respect to the methodology adopted for the project, the agile development method (extreme programming) was chosen for managing the project lifecycle. The reason for the decision was that, since it is a research intensive project it is better to develop it in relatively small iterations as more information becomes available. It would have been more risky for project completion, to have to wait until all required information for system design become available before proceeding to the implementation step as epitomised by the traditional waterfall methodology. This would have meant an increased project non-completion risk being taken. In addition, the time available for the project is limited, therefore the waterfall model being inherently inflexible would more likely have led to time over run, since any change in the requirements (as is common in a research project), would necessitate changes in several phases of the project leading to increased complexity (Zhang et al, 2010, p.174).

## 2    Literature Review

In this section different algorithms for object detection will be reviewed in an effort to justify the choice of the Local Binary Pattern classifier as the preferred algorithm for the Speed app. Some algorithms already applied to the detection of vehicle speeds will also be reviewed. It will be shown that the Speed app is a novel application of the Local Binary Pattern classifier for mobile speed detection for vehicles.

### 2.1    Computer Vision System Review

The Computer vision field focuses on facilitating a computer to understand or interpret visual information from images or video sequences and it originated in the late 1950s and early 1960s (Bebis et al., 2003, p.2). A typical vehicle surveillance system is composed of four stages: the foreground segmentation stage, the description stage, the classification stage and the tracking stage (Buch et al., 2011, p.923).

The foreground segmentation stage could be accomplished by calculating the difference between a background model and the current frame, as seen in (Gupte et al., 2002). It could also be achieved by computing the difference between two successive frames and thresholding the difference to create the foreground mask (Nguyen and Le, 2008). Another method for foreground estimation involves segmenting the foreground independent of the knowledge of the background. This could be realised by comparing the gradient image of the current frame to the gradient image of a 3D model wireframe. The foreground will then be computed by considering the match between the two images (Sullivan et al., 1997).

In the description stage, discriminative information or features are extracted from the foreground obtained previously. However, foreground segmentation is not always carried out, in such cases features could be obtained uniformly over the whole image, as is usually the case with features used with cascade classifiers. Another way of obtaining features is from the salient interest points in the image patch. Interest points are simply locations where features are extracted, while salient interest points may be defined as locations to obtain unique patterns that can identify an image, such as vertices of objects (Kapur and Thakkar, 2015, p.47). Popular salient interest point detectors are the Harris corner detector (Harris and Stephens, 1988) and the Features from Accelerated Segment Test (FAST) detector presented in (Rosten and Drummond, 2005).

The type of descriptor used determines if object scaling, translation, or rotation invariance, or any combination of them, will be a characteristic of the system. Some interest point descriptors which implement scale, translation, and rotation invariance are the Scale Invariant Feature Transform (SIFT) descriptor introduced in (Lowe, 1999) and the Speeded Up Robust Features (SURF) descriptor introduced in (Bay et al., 2006). The Speed app while being scale invariant is not invariant to rotation therefore the smartphone must be held in the proper orientation for effective detection.

In the classification stage of a traffic system an object with extracted features is mapped to an object class by an already trained classifier. A classifier is a trained model obtained as an output of a learning algorithm fed with a large set of training data or features, an example of such a learning algorithm is Adaptive Boosting (AdaBoost) (Freund, 1995). Some well-known classifiers are the Haar cascade classifier (Viola and Jones, 2004) and the Local Binary Patterns (LBP) cascade classifier (Ojala et al., 1996).

Tracking is the recognition of an object and its position between frames, it also involves constraints being imposed to mitigate noisy detector outputs by using a dynamic model. For tracking to be implemented, features have to be extracted in each frame for segmented regions and the extracted feature has to be matched in a consecutive frame to establish a consistent recognition of the tracked object.

## 2.2    Cascade Classifier Review

As mentioned in the Introduction, the Speed app has been developed using the OpenCV's implementation of the LBP in (Liao et al., 2007). The AdaBoost algorithm will be introduced first, then comparison will be made between the Haar cascade classifier and the LBP classifier.

Cascade classification in Computer Vision is typically accomplished using AdaBoost. During the learning step in the AdaBoost algorithm, a weak classifier or feature value with the lowest classification error is selected in any particular iteration from positive and negative samples. A classification error is the probability of a non-vehicle to be classified as a vehicle and a vehicle to be classified as a non-vehicle, while a positive sample is an image that contains a vehicle and a negative sample is a non-vehicle image. The learning stage terminates when a maximum number of features have been selected. The weak classifiers are then combined to form a strong classifier. Where there is a new image to classify, the image will be scanned with a fixed size pixel grid. (Viola and Jones, 2004, p.139) used a scanning window of 24 x 24 pixels for facial recognition but the Speed app uses a pixel grid of 48 x 24 pixels to accommodate the particular geometry of a vehicle.

Given that several segments of the new image will not contain a vehicle, a select number of weak classifiers are used in stages to determine if a window contains a vehicle or does not, based on a threshold. This improves the speed of the algorithm as a few features are used to eliminate non-vehicle windows in an early stage allowing computational power to be better focused on windows that are more likely to contain a vehicle in subsequent stages. (Viola and Jones, 2004, p.148) were able to reject 50% of non-face windows with just two features in the first stage of their cascade classification and also reject 80% percent of non-faces in the second stage of their classification with 10 features in a 38 layer cascade with a total of 6060 features. Next is the comparison between the Haar cascade classifier and the LBP classifier.

The Haar cascade classifier as presented by Viola and Jones (Viola and Jones, 2004) was a fast object detector. It achieved its speed by introducing the calculation of feature values with an integral image and by using cascaded AdaBoost classifiers. By using an AdaBoost classifier the algorithm is able select discriminant features from a pool of Haar features (Porwik and Lisowska, 2004) and by cascading detection speed is enhanced since non-matching windows of input images are quickly discarded in an early stage and detection can then be focused on fewer windows with less computing time.  The algorithm was first applied to face detection and it was 15 times faster than any previous method for face detection (Viola and Jones, 2004, p.152). It was considered a breakthrough in real time face detection (Zhang et al., 2007, p.11). The Haar cascade algorithm has also been applied to the detection of pedestrians in a traffic system (Jones and Snow, 2008).

Similarly, the Local Binary Patterns (LBP) cascade classifier uses the AdaBoost algorithm for feature selection (Huang et al., 2011, p771), however instead of Haar filters, it makes use of uniform patterns for feature extraction. The LBP was first proposed in (Ojala et al., 1996) and the pioneer use for face detection was in (Hadid et al., 2004). (Liao et al., 2007) extended the use of the LBP to compute feature values from multi-block sub-regions (MB-LBP) rather than computing from pixel values thereby capturing the macrostructures of patterns, they utilized integral images for faster computation and AdaBoost for feature selection. With a final strong classifier containing 2346 weak classifiers, (Liao et al., 2007) were able to achieve zero error rate on their training set because of the ability of the algorithm to encode both macro and microstructures. Their detection rate was 98.07% on a false acceptance rate of 0.1% on mask I, Experiment 1 data subset of the FRGC ver2.0 data set (Phillips et al., 2005).

The MB-LBP was compared to the Haar cascade classifier (Viola and Jones, 2004) by (Zhang et al., 2007) on the MIT+CMU database. 40,000 face samples were used to train the face detectors, the MB-LBP had 9 layers with 470 MB-LBP features while the Viola and Jones detector had 32 layers with 4297 Haar features. The MB-LBP classifier had a comparable performance to the Viola's detector while being more efficient. It is for this efficiency that the MB-LBP is preferable for object detection on mobile platforms and has been incorporated in the Speed app. In addition, in OpenCV, LBP feature value is an integer value against the Haar feature value which is a double value, therefore training and detection is much faster with LBP (Docs.opencv.org, 2016a).

## 2.3   Related Projects

Different projects have focused on detecting vehicle speeds from a single camera lens. (Ferrier et al., 1994) introduced the application of background approximation and difference to vehicle speed detection (Rad et al., 2010, p.2555). (Ginzburg et al., 2015) measured vehicle speeds using a computer and a mounted camera as a low cost alternative with less development time as compared to a typical traffic surveillance system. In their algorithm, after a foreground segmentation, based on a background subtraction sensitive to the colour of vehicle license number plate, they detected the location of the corner of the license plate of a vehicle across multiple frames and extracted the elapsed time from the frame rate and then computed the speed of the moving vehicles. Vehicle tracking was accomplished by classifying the earlier segmented region as a licence plate using the Support Vector Machines (SVM) classifier (Vapnik, 1982) and extracting the licence number using a three layer Artificial Neural Network (ANN) classifier. The capturing camera was placed on the side of road surface to oversee the incoming traffic. Due to the distortion in their captured image they computed a homographic matrix to handle the image plane projection on the road surface. They also introduced a speed correction factor based on the height of the vehicle above the road surface. Their speed detection varied proportionally to the ground-truth speed obtained by a GPS speedometer.

Their speed correction factor is difficult to implement because of the need to have a prior knowledge of the height of the moving vehicles above a road surface and is not featured in the Speed app. Furthermore, the factor may be negligible since the differences in the height of vehicles above a road surface may not on average be significant. Another disadvantage of the (Ginzburg et al., 2015) method is having to know the licence number plate colour in advance. This could be avoided by detecting the whole vehicle using an AdaBoost cascade classifier and then scanning the segmented region for the number plate surface using a Canny edge detector, and using Hough transform for line detection and iterating through the detected lines to determine the corners for the bounding rectangle of the license plate, which is limited to threshold size. Number plate recognition has not been incorporated into the Speed app given that speed is measured from the side view of a vehicle.

(Ginzburg et al., 2015) employed homography because the camera lens was meant to be placed above the road surface with its optical axis tilting downwards towards the road section's forward direction. By contrast the Speed app does not employ a plane transformation matrix because it captures the side view of a moving vehicle and not its overhead view. The side view is already parallel to the camera's perpendicular field of view. Also, the Speed app detects the LBP features of a vehicle and not the license plates' corners in tracking the location of a vehicle. In addition the Speed app could be a lower cost system since it does not require separate systems to implement vehicle speed detection, however it is not suited for the same application. The system developed by (Ginzburg et al., 2015) was designed to be a stationary overhead speed camera while the Speed app is designed to be a pedestrian speed camera hence the difference in camera angles.

Another vehicle speed detecting system was proposed by (Rad et al., 2010). They used a video camera, a computer and MatLab software to implement their system, still as a low cost alternative and a more reliable system to radar surveillance traffic systems. Background estimation was accomplished by using the mean filter method and subtraction was achieved by using the combined saturation and value method (CVS). Edge detection and morphological operations were applied to the binarised image to segment the vehicles in the image and find their bounding boxes. The location of the moving vehicles centroids were then tracked and with the image frame rate, the vehicle speeds were computed. To calibrate their camera to record real speeds, they compared their image height to the height of the detection area and obtained a resolution, they further computed a calibration coefficient which was derived by tuning the detected speed to an actual vehicle speed known in advance in an effort to improve the accuracy of their system. They were able to achieve correlating speeds with an average error of ± 7km/h. The camera for the (Rad et al., 2010) system was required to be positioned above the roadway with its optical axis tilting towards the roadway, which is in contrast to the intended use of the Speed app.

The speed detecting system of (Rad et al., 2010) was reliable and like the work of (Ginzburg et al., 2015) discussed previously, it was designed to serve as a stationary overhead speed camera. The system demonstrated that the problems of a typical radar speed camera could be avoided with a low cost alternative. The problems of a radar traffic surveillance camera typically include having high cost, radio interference where they are radio signals of similar frequency in the detection area, dealing with cosine error, where the axis of an incoming vehicle is not aligned to the axis of the radar system, and having to track only one vehicle at a time, on the contrary, a computer vision system can track the speeds of multiple vehicles at any particular time. The (Rad et al., 2010) system was a multi vehicle tracking system.

Similarly, the vehicle speed detection system proposed by (Wu et al., 2009, p.196) is conceptually identical to that of (Rad et al., 2010) but did not incorporate the calibration coefficient and they achieved an average error was 3.3km/h.

All the systems described above were designed to be fixed overhead speed camera systems, however, a speed camera on a mobile device would provide greater comfort for the user. The multi vehicle speed detection accomplished mainly by feature extraction of foreground segments could still be implemented on a mobile device however with respect to OpenCV, it would require calling the OpenCV native C++ API to be able to implement such a system in real-time. The multi vehicle speed detection feature has not been implemented in the Speed app which uses the OpenCV Java API, but it has been deferred for future developments on the project.

So far, no vehicle speed detection system on a mobile smartphone, using computer vision has been encountered by the author. It is the intent of this report to fill that gap and suggest practical ways by which such a system can be further developed. The Speed app goes further than just computing vehicle speeds to the management of the data obtained in the process of surveillance. The facility to upload accumulated data in the Speed app to a webserver has also been included in the proposed system.

## 2.4   Legal Social Ethical and Professional Issues

They are laws and guidelines that need to inform the legal, social, ethical and professional aspects of the presented Speed app project. Such laws are: the Data Protection Act, Telecommunications Act 1984 and the Computer Misuse Act 1990. The Data Protection Act requires that personal data in ordinary circumstances should be obtained only with the consent of the subject or for legitimate interest without prejudice or in the case of sensitive personal data, be obtained with appropriate safeguards to protect the right to privacy in addition to consent (Legislation.gov.uk, 2016a). In keeping with the Data Protection Act, the app test has been carried out only with the consent of the subjects and portions of the project images capturing passer-by's have been blurred out. Furthermore in line with both the professional Code of Conduct for BCS Members and the Universal Ethical Code for Scientists (British Computer Society, 2015; Government Office for Science, 2007) the test results of the project have been reported accurately and objectively.

## 3    App Analysis and Design

The project requirements will be analysed and represented using appropriate Unified Modelling Language (UML) diagrams. The UML diagrams of the system that will be considered will be the use case diagram, the class and sequence diagrams for the Detect vehicle speed use case, which is the major use case of the system. The entity relationship diagram (ERD) which is usually modelled to reflect the database design of a software system will not be considered because it is trivial with respect to the project, since the Speed app SQLite database contains only one table.

The implementation class diagram of the project and the detailed sequence diagram for key methods implementing use cases in the project have been included in the Appendices for convenience. The UML diagrams have been developed using Visual Paradigm Professional Edition Version 12.2.

### 3.1    Use Case Analysis

The use case analysis will be presented both specified in writing and graphically. From the objectives of the project as stated in the Introduction, the following use cases were deduced:

*Table 3.1: Use case specification*

| Actor | Use Case |
|---|---|
| Traffic Officer | Detect vehicle speed |
| | Obtain location information |
| | Capture detection time |
| | Take vehicle picture |
| | Save generated data |
| | View saved data |
| | Search saved data by date field |
| | Delete saved data |
| | Upload saved data to web server |

The use cases in Table 3.1 above will be illustrated graphically in Figure 3.1 below to allow for more insight into the relationships existing in the system.

In Figure 3.1, the Traffic officer is shown as the sole actor in the system. There is an include relationship between the Save generated data use case and the Detect vehicle speed, Obtain location information, Capture detection time, and the Take vehicle picture use cases because logically the later use cases need to be completed before the Save generated data use case can be executed.

However, the include relationship is not direct, it is being implemented through the Capture data use case. The Capture data use case is a generalisation relationship for the Detect vehicle speed, Obtain location information, Capture detection time, and Take vehicle picture use cases, it is included for convenience in the diagram but is not included in the table of use cases. Conversely, the View saved data use case, Search saved data by date field, Delete saved data, and the Upload saved data to web server use cases,  should only be only be executed after generated data in the app have been saved, this explains the include relationship between the use cases and the Save generated data use case.

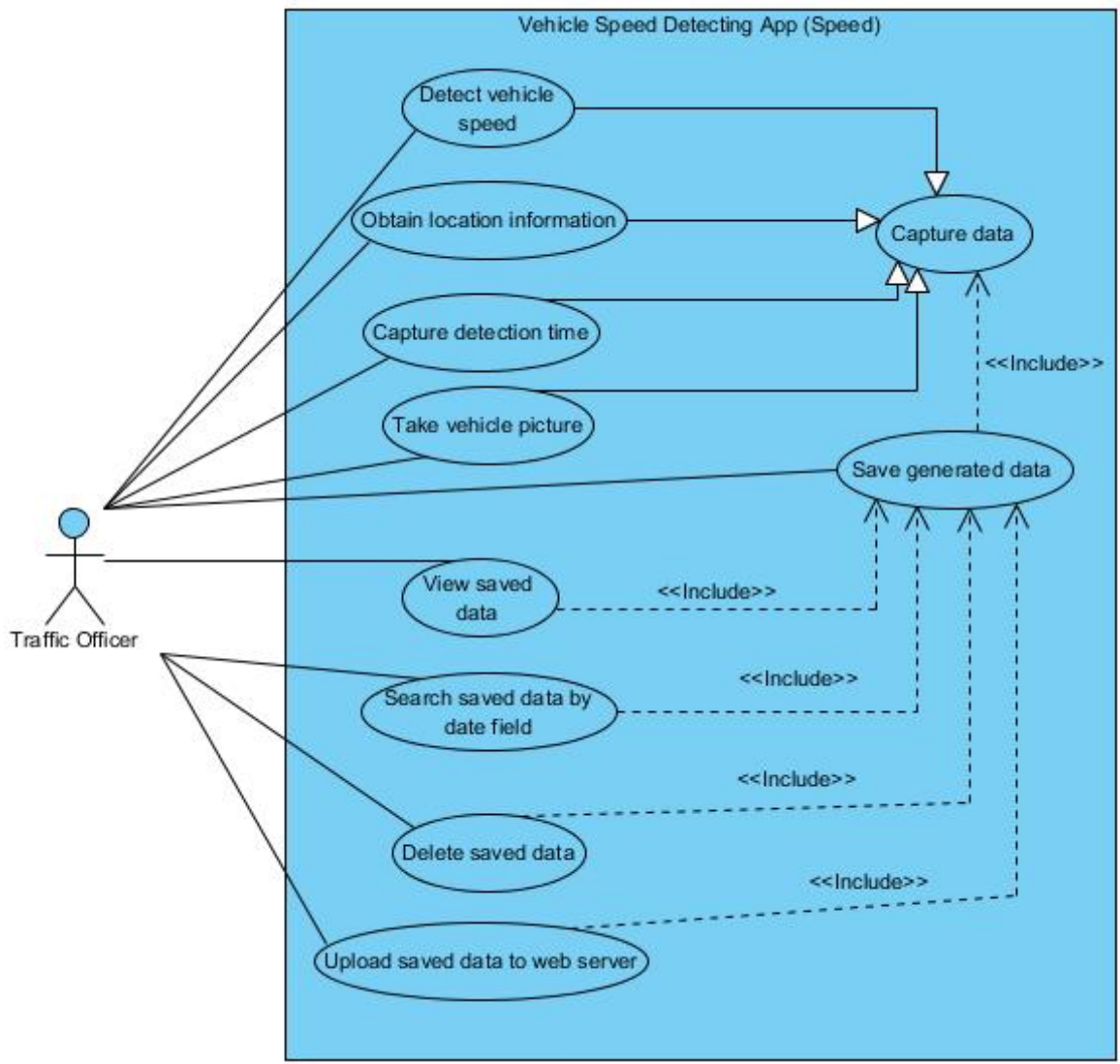

*Figure 3.1: The system use case diagram*

## 3.2 Class Diagram

The class diagram shown in Figure 3.2 below contains the Activities and other classes involved in the Detect vehicle speed use case. The detailed project class diagram showing the inter relationship between the classes has been included in the Appendices. In Android parlance, a class that contains the source code to be run for a visual form, interface, or page to be created and made functional is called an Activity (Activities | Android Developers, 2016).

Figure 3.2 shows the three Activities involved in the Detect vehicle speed use case of the project. The classes without a package name all belong to the default com.project.itoro.speedapp package. The plus symbol (+) represent public elements, the minus symbol represents the private elements, while the hash symbol (#) represents the protected elements of the Activities.

The filled arrows represent the composition relationship between classes, while the unfilled arrows represent the aggregation between the classes. The aggregation relationships occur as a result of the conscious effort to modularise the project and be object oriented. Codes that otherwise would have been bumped into the life cycle methods of Activities, as is common in mobile development, were analysed and where feasible grouped into classes to allow for better maintainability and to conform to the object oriented programming principle of high cohesion, in this case, separating implementation codes from program logic. The labels beside the relationship arrows are variable names that create the connection.

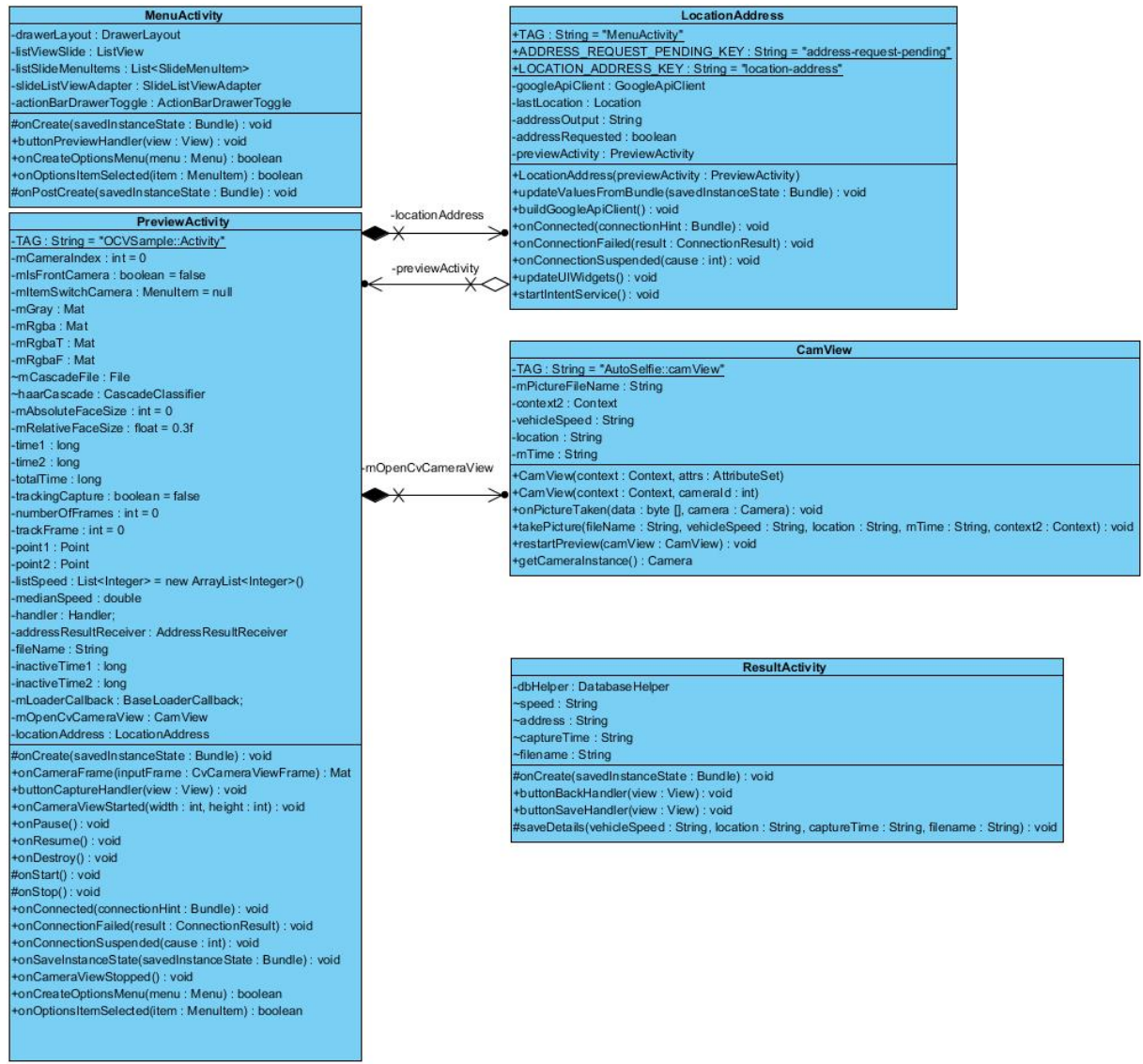

*Figure 3.2: The Activity class diagram for the Detect vehicle speed use case*

## 3.3  Sequence Diagram

The sequence diagram for the Detect vehicle speed use case is shown in Figure 3.3 below. From the diagram the Traffic officer is the actor. The Speed app opens to the MenuActivity as the home screen, on the loading of the MenuActivity the lifecycle methods onCreate(), onCreateOptionsMenu() and onPostCreate() will be executed by default. The onCreate method handles the creation of the user interface (UI) elements, the onCreateOptionsMenu handles the population of the action bar (i.e. the menu) and the onPost method serves to map the state of the menu icon to the state of the drawer layout. The self-message symbol used for the life cycle and callback methods is to indicate that the methods are being triggered from within the owner objects by Android. Anonymous objects have been used in the creation of the sequence diagram so as not to make the diagram more complex with variable names.

When a user clicks the Preview button on the MenuActivity the PreviewActivity will be launched. The PreviewActivity launch is being carried out from the buttonPreviewHandler() method which handles the Preview button click event. The messages in the sequence diagram without brackets are not actual method names but description of user actions, to make the sequence diagram more readable.

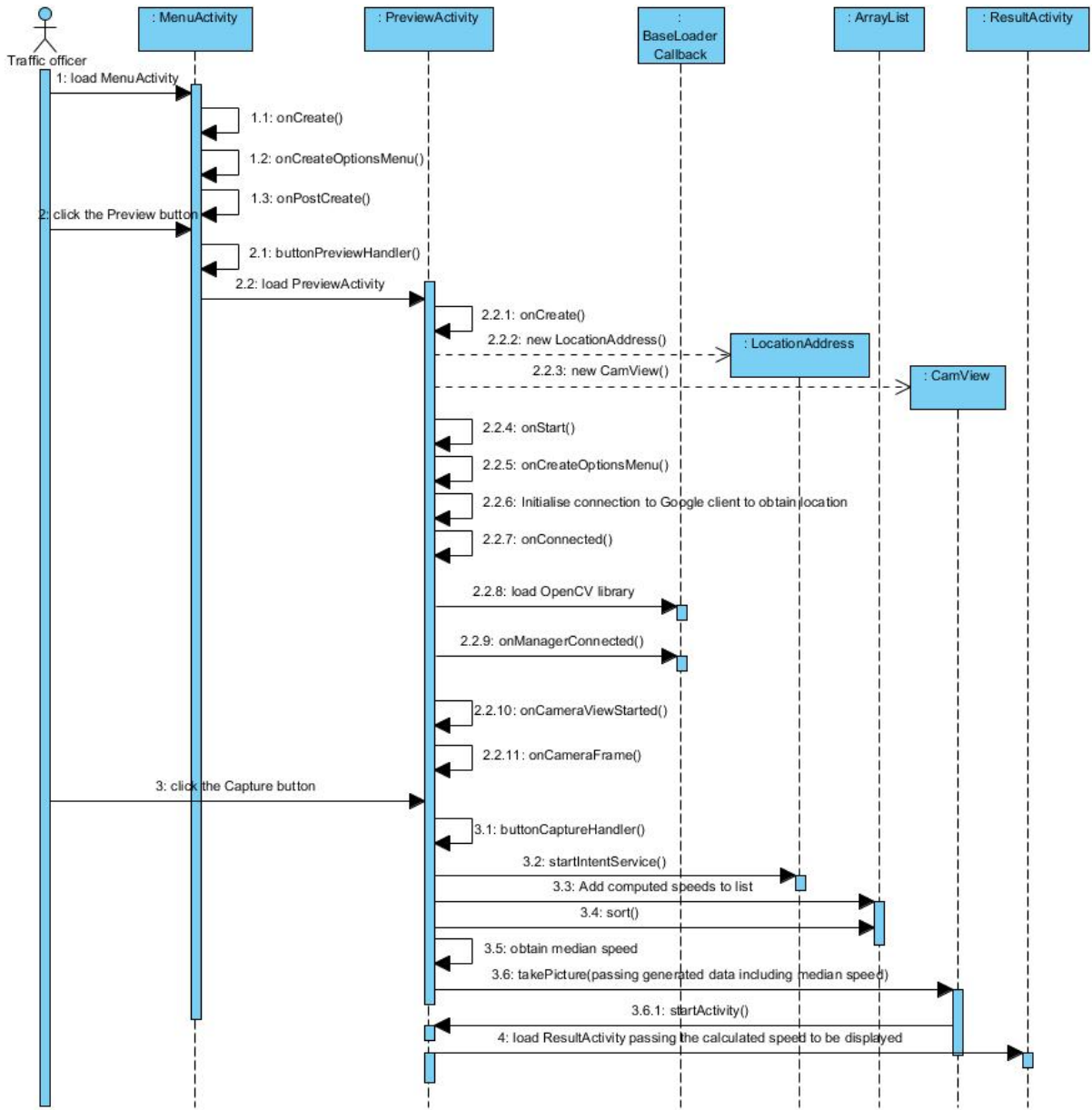

*Figure 3.3: The sequence diagram for the Detect vehicle speed use case*

The PreviewActivity loads with its creation lifecycle methods running and are indicated with the self messages. The LocationAddress and the CamView objects have been so positioned to indicate that they are variables local to the onCreate method. The BaseLoaderCallback object connects the app with the OpenCV library. The OpenCV library comes as a separate application that has to be installed on a mobile device that runs an app that calls the library. Therefore the OpenCV's BaseLoaderCallback object bridges the two applications. When OpenCv has been successfully connected to the Speed app, the onManagerConnected() callback method of the BaseLoaderCallback object is triggered, making the method the ideal place to instantiate

OpenCV related objects. It is in the onManagerConnected() method that the LBP cascade classifier is instantiated. The BaseLoaderCallback object is defined inside PreviewActivity.

The PreviewActivity UI contains the OpenCV camera view. In Android, a view is a place holder on an Android user interface. It is the fundamental building block for any UI element (View | Android Developers, 2016). For frames captured by the smart phone camera to be displayed on screen, there must be a camera surface view to display the captured image frames. So, once OpenCV is loaded the OpenCV JavaCameraView will then begin to display the frames captured by the smartphone camera. Later in the Implementation section it will be further explained how the camera previewing was actually carried out. Before any frame is displayed by the JavaCameraView the onCameraFrame() method must first be called. The onCameraFrame() method is where image processing is usually carried out before being displayed on device screen.

When the Capture button is tapped on the PreviewActivity, the button's handler, the buttonCaptureHandler() method is executed. This is followed by the startIntentService() method which fetches location information. The buttonCaptureHandler() method sets the value of a certain variable Boolean variable "trackingCapture" to true. The trackingCapture variable tracks if the Capture button has been clicked. Where trackingCapture is true, the calculation of vehicle speed will commence in the onCameraFrame() method. When a speed value is obtained, it is passed along side location and time information to the takePicture() method of the CamView class, where they will be further passed to the ResultActivity to be displayed. Since the sequence diagram is for the Detect vehicle speed use case, details on the capturing of other data, such as picture data will be deferred to the Implementation section of this report.

## 4   Implementation

The implementation of the full scope of the project will be demonstrated with appropriate screenshots and with the rationale for implementation decisions being discussed. The discussion will follow a serial approach commencing from when a user opens the app. Before then, the justification for choosing Eclipse IDE Version: Mars.2 Release (4.5.2) as the preferred development environment for the app will be considered.

### 4.1   Justification for Choice of IDE

There are two major platforms for Android app development, the Android Studio IDE and the Eclipse IDE. Eclipse is the former official platform for Android development while Android Studio is now the official Android development IDE (Meet Android Studio | Android Studio, 2016) and was introduced by Google in 2013 (Android Studio: An IDE built for Android, 2013). Google's support for Eclipse ended in 2015 and Android Studio was promoted by Google as a powerful development suite. However in the author's experience in using the two IDEs, Eclipse performed at least 10 times faster than Android Studio in project building on a 4GB RAM computer with Intel N3050 processor at 1.6GHz. In the time critical Speed app project, Eclipse was inevitably chosen as the preferred IDE for development.

### 4.2   App Development

Figure 4.1 below shows the app icon for the application, it was designed using the Adobe Photoshop CC 2015.5.0 Release and was integrated into the app using the Android Icon Set wizard of Eclipse IDE.

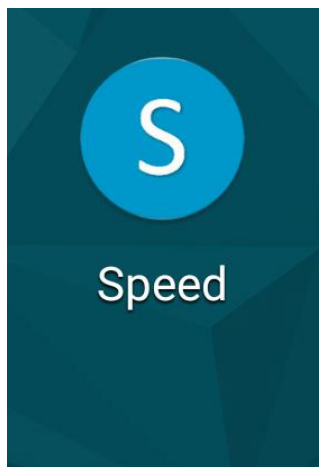

*Figure 4.1: The Speed app icon*

Clicking the icon opens the MenuActivity of the app. The screenshot of the MenuActivity is shown below in Figure 4.2. The layout of the MenuActivity has been designed using an image from the Internet (Pixabay.com, 2016) and a custom designed Preview button. The circular shape of the button was achieved by making appropriate adjustments to the background property of the button in its layout file. The background property was made a reference to the custom_color_button.xml drawable file. In the drawable file, the <selector> element and <item> element having a <shape> child element were used to define the circular shape of the button. Alternatively, a <bitmap> child element could have been used with a circular drawable bitmap to still accomplish the same result. Figure 4.3 below shows the content of the custom_color_button.xml file.

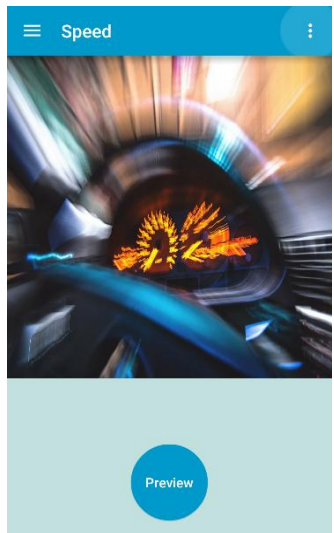

*Figure 4.2: The MenuActivity layout*

```
custom_color_button.xml ⊠
1  <?xml version="1.0" encoding="utf-8"?>
2  <selector xmlns:android="http://schemas.android.com/apk/res/android">
3      <item>
4          <shape android:shape="oval">
5              <stroke android:color="@color/colorPrimary" android:width="5dp" />
6              <solid android:color="@color/colorPrimary"/>
7              <size android:width="100dp" android:height="100dp"/>
8          </shape>
9      </item>
10 </selector>
```

*Figure 4.3: The content of custom_color_button.xml file*

Figure 4.3 below shows the MenuActivity with an opened drawer layout. The drawer layout is a common feature of Android template apps. Icons have been used alongside the menu options as is obtainable in standard Android UI design templates.

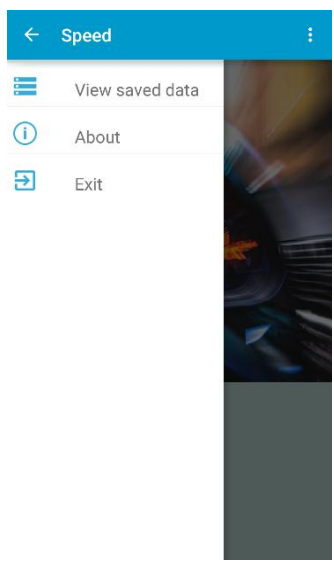

*Figure 4.4: The MenuActivity layout showing menu on a drawer layout*

On clicking the About option in the sliding drawer the about screen of the app is displayed as shown in Figure 4.5 below.

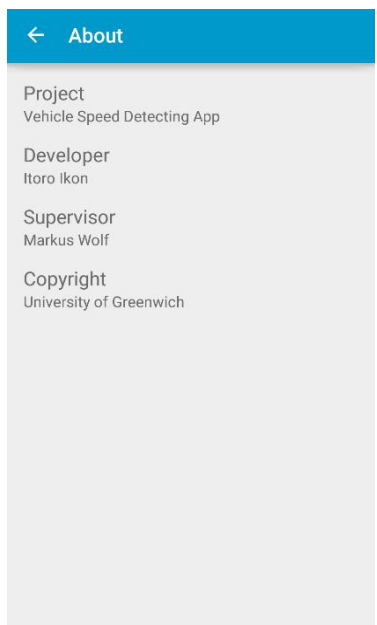

*Figure 4.5: The about app screen*

When the Preview button is clicked in the MenuActivity, the PreviewActivity will be loaded as shown in Figure 4.4. The PreviewActivity is pictured with the smart phone camera lens focused on the image (Borongaja, 2016) of a car on a screen. As seen in Figure 4.4, when the app detects the image of a vehicle, a bounding rectangle will be formed around the detected vehicle. The author did set the bounding rectangle to a minimum 30% of the display screen (through the mRelativeVehicleSize variable being equal to 0.3f in PreviewActivity), this coincides with the minimum area in which OpenCV will scan for vehicles. If the value is lower more computation power will be required as more scans will be needed per frame and the probability of false detections will be higher. With a higher value, false detections are minimised, but if too high will result in no detections being made. So it is one of the parameters that determine the sensitivity to detection and was set after some trials.

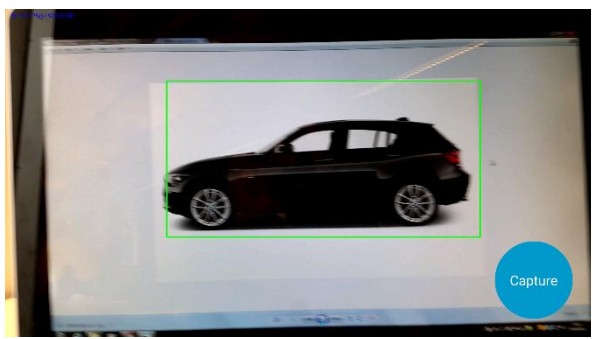

*Figure 4.6: The PreviewActivity screen showing a detected vehicle*

Figure 4.7 below shows the instantiation of the main engine of the project; the LBP cascade classifier (Kapur and Thakkar, 2015, p.87). The snippet is from the onManagerConnected() callback method of the BaseLoaderCallback object in the PreviewActivity. The trained cascade classifier model is saved in the cascade.xml file. To train the classifier for the project the author used images provided in the database of the Cognitive Computation Group (Cogcomp.cs.illinois.edu, 2016). The OpenCV 2.4.11 for Windows (Sourceforge.net, 2016) was used together with the images to train the classifier (Tutorial | packtpub.com, 2013). The classifier training used 550 positive samples and 500 negative samples and completed 17 training stages in about 5 hours on a 4GB RAM computer with Intel N3050 processor at 1.6GHz. The code below reads the cascade file from the project's raw folder to the internal storage of the mobile device for access and then the OpenCV CascadeClassifier class is instantiated with the cascade.xml file being passed to the constructor.

```java
try
{
    InputStream is = getResources().openRawResource(R.raw.cascade);
    File cascadeDir = getDir("cascade",Context.MODE_PRIVATE);
    mCascadeFile = new File(cascadeDir, "cascade.xml");
    FileOutputStream os = new FileOutputStream(mCascadeFile);
    byte[] buffer = new byte[4096];
    int bytesRead;
    while((bytesRead = is.read(buffer)) != -1)
    {
        os.write(buffer, 0, bytesRead);
    }
    is.close();
    os.close();
    lbpCascade = new CascadeClassifier(mCascadeFile.getAbsolutePath());
    if (lbpCascade.empty())
    {
        Log.i("Cascade Error","Failed to load cascade classifier");
        lbpCascade = null;
    }
}
catch(Exception e)
{
    Log.i("Cascade Error: ","Cascase not found");
}
```

*Figure 4.7: The instantiation of the LBP cascade classifier*

With respect to the design of the visual interface of the PreviewActivity, the FrameLayout view and a button view were utilised. As was mentioned earlier when discussing the sequence diagram for the Detect vehicle speed use case, for a camera frame to be displayed on screen it needs to be coupled to a surface view. OpenCV provides the JavaCameraView class to serve this purpose. However, the JavaCameraView was not directly used to display the frames, rather it was added to the FrameLayout for the PreviewActivity. The FrameLayout was chosen to so that it will be possible to place the Capture button view on top of the camera frame display, while the displayed frame will still have a resolution matching the dimensions of the FrameLayout.

Furthermore, the JavaCameraView provided by OpenCV was not directly added to the FrameLayout but was first extended by the CamView class, and a CamView object was then added to the FrameLayout. The extension was necessary so that there could be access to the underlying Android Camera object. The Android Camera object needs to be accessed to set a parameter like frame rate. An alternative to the JavaCameraView is the NativeCameraView which could have more stable frame rates, it is provided by OpenCV but it is not supported on all Android devices and was not supported on the mobile device used to test run the app. The mobile device used in test running the Speed app is the Samsung Galaxy S5 (SM-G900F) running Android version 5.0 API level 21.

On the top left corner of the PreviewActivity, the OpenCV's current frame rate and image resolution are shown. The resolution does not always match the dimension of the displaying view but OpenCV will choose a supported image resolution closest to the dimension of a displaying view. For the Speed app the frame resolution is 1920 x 1080. Although the frame rate for the Speed app has been set at 30 frames per second (fps), frame rates are not guaranteed by Android and could still fluctuate depending on the prevailing light conditions. This informed minimising the number of frames used for detecting vehicular speed.

On clicking the Capture button, the coordinate of the top left corner of the bounding rectangle is tracked across 20 frames. After each set of five frames in which a vehicle has been detected, the Euclidean distance is calculated between the initial and final top left corner coordinate of the bounding rectangle. OpenCV sets the coordinates of a camera surface such that the top left corner of an image surface is coordinate (0,0) while the bottom right corner is the maximum on both the horizontal (x) and vertical (y) axis. Figure 4.8 shows the source code from the onCameraFrame() method for speed calculation after five frames.

```java
//Calculating speed for every five frames of detection
if(trackFrame==5)
{
    time2 = System.currentTimeMillis();
    point2 = vehicleArray[0].tl();

    double xDiff = point2.x - point1.x;
    double yDiff = point2.y - point1.y;

    double dist = Math.sqrt((xDiff * xDiff) + (yDiff * yDiff));

    double speed = 0.25*((dist)/((double)(time2 - time1)))*1000.0;

    listSpeed.add((int)Math.round(speed));

    trackFrame = 0;
}
```

*Figure 4.8: Code snippet showing speed calculation*

The int variable trackFrame counts the number of detections, the long variable time2 stores the android device's time in milliseconds in the long integer format at the end of the five detections, time1 is similar to time2 except that it stored the initial system time. point2 is an object of the OpenCV Point class which stores the top left coordinate of the final bounding rectangle, while point1 stores the initial coordinate. The scalar value 1000.0 converts the calculated speed to pixels per second. The 0.25 is an arbitrary coefficient to map the calculated speed to the ground truth speed. The final coefficient value and the conversion from pixels per second to kilometres per hour will be discussed on the System Testing section.

On reaching 20 detections, the computed speeds are sorted and the median speed is selected as the optimal vehicle speed. The median speed was preferable to cushion against a situation where a false detection is encountered, the computed speed value for a false detection will likely be further away from the speed values obtained from true detections. A median value ensures that a speed farther away from the central tendency speed is not selected and does not influence the selected speed. An alternative could be to use the mean value but the disadvantage is that a mean value will be skewed to the magnitude of the speed from a false detection while a median speed is better positioned to eliminate the effect of false detections, a similar method was used by (Ginzburg et al., 2015).

With the speed calculation concluded, other data will also be captured. These are the current system time, the geographical location for speed capturing, the picture of the speeding vehicle and also a file name being generated for the captured picture. The current system time is generated from the Java Date and SimpleDateFormat classes. The geographical location was obtained using the Google Play services API. The current version of the API client is not supported in Eclipse but this problem was avoided by using an older version obtained from Google repository at (Google repository, 2016). The vehicle picture taking was accomplished by calling the takePicture() method on the Android Camera object. And picture filename was generated from the getPath() method of the Android File class. The current system time was appended to the picture filename to avoid potential conflicts in filenames on the Android device.

Having obtained the generated data the ResultActivity class was called to display the captured information. Initially the Android AlertDialog message box was preferred to display the generated data due to the simplicity of its use, but a missing Handler error was encountered for attempting to display the AlertDialog from the onCameraFrame() method. To solve this problem the ResultActivity was created to display the generated data. However, on attempting to start the ResultActivity from the onCameraFrame() method, the startActivity() command would not execute and no errors were thrown by the IDE. With the knowledge that after a takePicture() method is called on an Android Camera object then the onPictureTaken() callback method will be called, the startActivity() method was moved to the Camera callback method were it finally executed. The onPictureTaken() callback method is usually used to save the bytes obtained from the camera frame to storage.

The picture taken when measuring speed is saved to the Android internal storage (Context.MODE_PRIVATE). Another option could be to save the picture to the external storage. Saving to the internal storage was preferred because there will be no access to the saved file except by the app therefore enhancing the security of the snapped picture against unauthorised access. The other generated textual information were saved to the incorporated SQLite database including the saved picture's filename. The picture filename in the SQLite database is later used to retrieve the picture file from device storage for uploading to the web server. The SQLite database is a lightweight database designed to be used in mobile application development.

Figure 4.9 shows the ResultActivity screen as a dialog box. This was achieved by setting the android:theme property of the Activity to Theme.Dialog. The ResultActivity shows the vehicle speed, the location of capture, the time of capture and the snapped picture filename. If the Back button is pressed the data are discarded but if the save button is pressed the generated data will be saved as described above, the ResultActivity will then be destroyed while the PreviewActivity will be resumed.

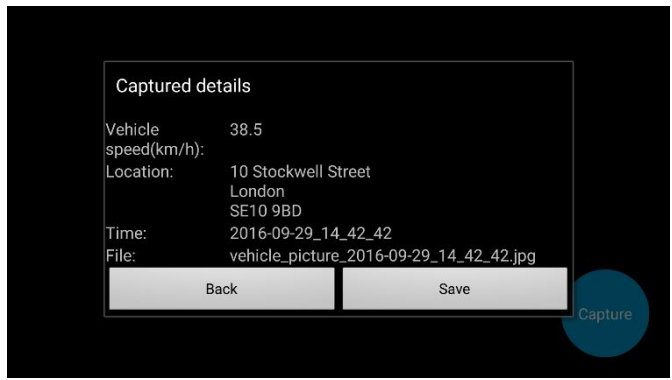

*Figure 4.9: The ResultActivity screen being shown as a dialog*

To view the data saved by the app, the Details Activity needs to be started through the View saved data menu on the drawer list as shown in Figure 4.4. In Figure 4.10 below, all the accumulated data in the Speed app are displayed on a list. A list is preferable in displaying the data since it allows scrolling.

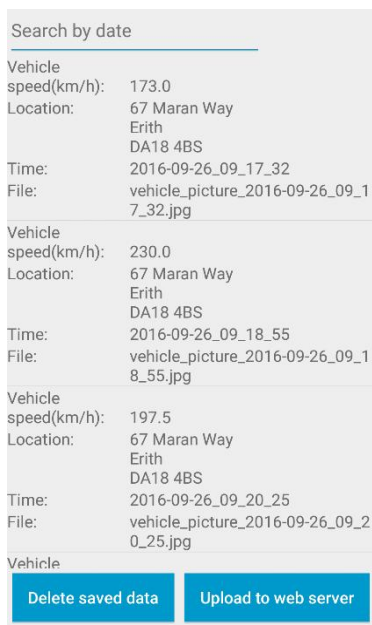

*Figure 4.10: The Details Activity showing all saved data*

The Detail Activity also contains a Search by text EditText view. Since the list is designed to display a large amount of data it makes sense to provide a facility that will enable quick access to any desired data. The EditText view allows the list to be searched on its Time field. The MTextWatcher class extending Android TextWatcher class listens for text change event on the EditText view and filters the list through its onTextChanged() callback method.

The two buttons on the Detail Activity are for deleting accumulated data and for sending accumulated data to web server respectively. Clicking the Delete saved data button will bring up the AlertDialog asking for confirmation as shown in Figure 4.11.

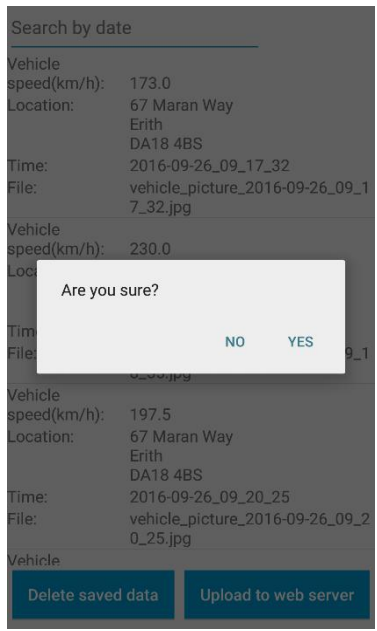

*Figure 4.11: The AlertDialog for confirmation before deleting saved data*

On the contrary, clicking the Upload to web server button will transfer all saved data to the web server through an exposed web service. This is designed to mirror the situation where a Traffic officer needs to transfer information on his mobile to a central data centre.

The uploading of data to the web server was implemented through the Java URL and HttpURLConnection objects. The HTTP POST method was employed to transport the device's data to web server. The alternative GET method would not have been feasible because the relatively large amount of data to be transferred to the web server could not be handled by the GET method due to server constraints. For example during app test it was realised that the maximum value allowed for the maxUrlLength property of the IIS web server of Microsoft Azure cloud service was only 2097151 characters, which cannot practically allow the uploading of picture files.

The saved data was transferred as a JSON string. However, since picture bytes can include characters that are not acceptable by JSON standards (JSON, 2016), the Base64 encoding algorithm (Base64, 2016) was used to convert picture bytes into Base64 strings before being eventually converted to JSON string. The forward slash (/) character of the Base64 algorithm was replaced by the underscore character (_) because slashes cannot be directly transported by HTTP without URL encoding. On the server side the JSON string is decoded to a class and the Base64 string is decoded back to picture bytes to be saved in the filesystem provided in the Microsoft Azure cloud service.

The codes to send the JSON data and the interface code on the server side were developed by the author as there were no explicit code snippets showing how to make a post request to a Windows Communication Foundation (WCF) web service in C# from android platform. The codes were developed by analogy from studying codes in Ajax for making HTTP POST call to a C# WCF service receiving XML data. All efforts on searching the Internet and placing calls to colleagues proved abortive in finding an identical solution and hence having to rely on personal effort. Figure 4.12 shows the connection setting required to make the HTTP client POST call from Android, found in the JsonActivity class, while Figure 4.13 shows the WCF service interface for accepting a JSON POST request. The con variable is the HttpURLConnection object.

```
try
{
    // configure the connection to do a POST
    //Set the DoOutput flag to true if you intend to use the
    //URL connection for output
    con.setDoOutput(true);
    con.setRequestMethod("POST");
    con.setRequestProperty("Content-Type", "application/json");
}
catch (Exception e)
{
    e.printStackTrace();
}
```

*Figure 4.12: Setting the HttpURLConnection object for a POST request*

```
[ServiceContract]
public interface IProjWcfService
{
    [OperationContract]
    [WebInvoke(Method = "POST",
        UriTemplate = "upLoadData",
        RequestFormat = WebMessageFormat.Json,
        ResponseFormat = WebMessageFormat.Json,
        BodyStyle = WebMessageBodyStyle.Bare)]
    UploadResponse upLoadData(UploadData jsonString);
}
```

*Figure 4.13: The WCF service interface for accepting a JSON POST request*

In the implementation of the web server side of the Speed app project, the Windows Communication Foundation (WCF) web service (Windows Communication Foundation, 2016) was chosen as the preferred web service for the project. The WCF is the Microsoft Service Oriented Architecture (SOA) framework for developing service oriented applications. An alternative to the WCF is the ASMX web service. The ASMX web service is becoming a legacy technology because of its security limitations and lack of robustness, for example it can only be run on HTTP while WCF can run on other transport/application layer protocols (ASMX and WCF Services, 2016).

Furthermore, the WCF service implemented in C# was used in the project rather than Java web services because an application that can run on the Microsoft IIS web server can be readily hosted on the University's IIS servers. However, the Microsoft Azure cloud service was finally chosen rather than the University's server, because of the security challenges the University faced with its servers at the initiation of the project. The Microsoft Azure is the cloud service of Microsoft and offers a range of services such as web hosting, SQL database service, filesystem service which integrate seamlessly with Microsoft Visual Studio and Microsoft SQL Server Management Studio.

Microsoft Visual Studio Professional 2013 Version 12.0.40629.00 Update 5 and Microsoft SQL Server Management Studio 13.0.15000.23 were used to implement the WCF web service and the web server SQL database respectively before they were deployed to MS Azure.

Figure 4.14 shows the message returned by the WCF web service to the app's JsonActivity acknowledging the reception of records. The message has been displayed on the Activity's WebView.

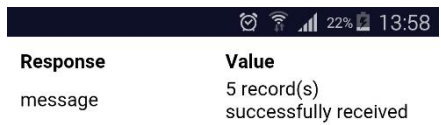

*Figure 4.14: The JsonActivity showing successful upload of data to web server*

Figure 4.15 shows the uploaded image files being saved in the Microsoft Azure image Storage account as seen from Google Chrome browser, while Figure 4.16 shows the content of the SQL database deployed on Azure after data was successfully uploaded. In line with best practices, the image file was stored separately in Azure storage account while textual information was saved in SQL database. This allows for easier viewing of the picture files than if they were stored as blobs or bytes in the database. The Microsoft SQL Server Management Studio was logged into with Azure credentials to be able to view the data stored in the Azure SQL database as seen in Figure 4.16.

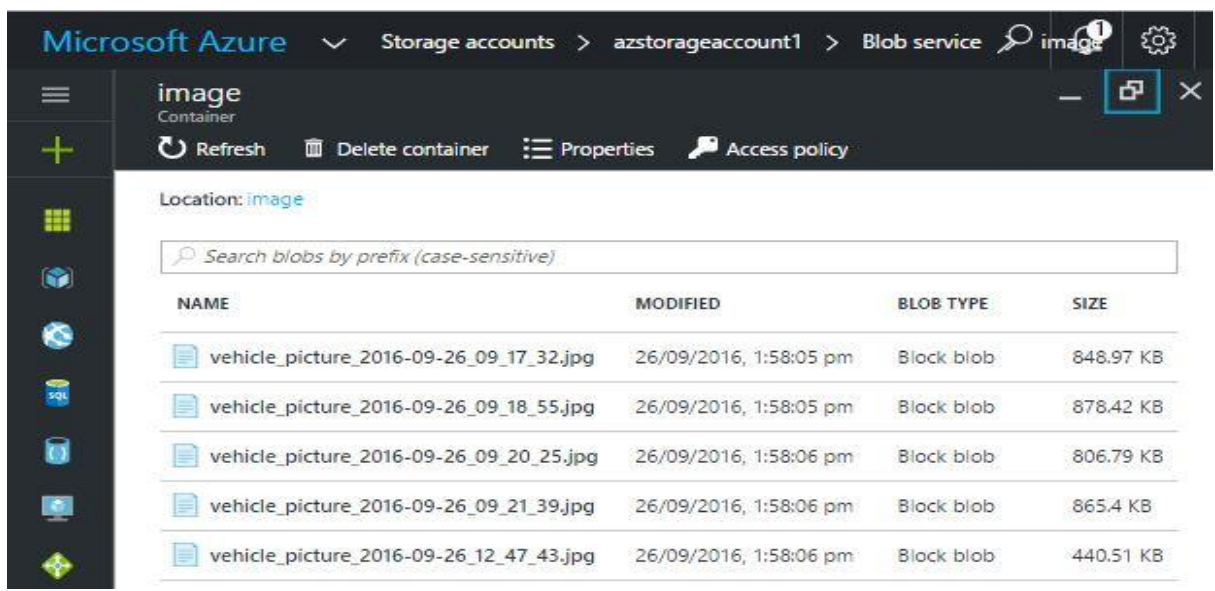

*Figure 4.15: The uploaded image files shown in Microsoft Azure*

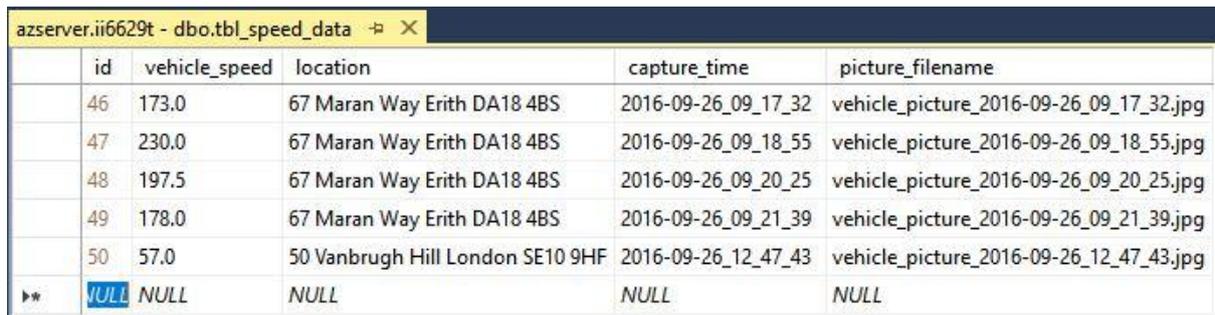

| | id | vehicle_speed | location | capture_time | picture_filename |
|---|---|---|---|---|---|
| | 46 | 173.0 | 67 Maran Way Erith DA18 4BS | 2016-09-26_09_17_32 | vehicle_picture_2016-09-26_09_17_32.jpg |
| | 47 | 230.0 | 67 Maran Way Erith DA18 4BS | 2016-09-26_09_18_55 | vehicle_picture_2016-09-26_09_18_55.jpg |
| | 48 | 197.5 | 67 Maran Way Erith DA18 4BS | 2016-09-26_09_20_25 | vehicle_picture_2016-09-26_09_20_25.jpg |
| | 49 | 178.0 | 67 Maran Way Erith DA18 4BS | 2016-09-26_09_21_39 | vehicle_picture_2016-09-26_09_21_39.jpg |
| | 50 | 57.0 | 50 Vanbrugh Hill London SE10 9HF | 2016-09-26_12_47_43 | vehicle_picture_2016-09-26_12_47_43.jpg |
| ▶* | NULL | NULL | NULL | NULL | NULL |

*Figure 4.16: Textual data being saved to SQL database deployed on Azure*

## 5   System Testing

The Speed app was tested twice. The test was conducted by keeping the mobile device running the Speed app at a fixed distance from the side of the road and by extension the vehicle for testing. After the first test it was necessary to adjust the number of required vehicle detected frames from an initial value of 100 to 20. The vehicle used for testing was going off the screen before 100 detections could be obtained. With 20 required frames this problem was solved. Figure 5.1 shows a picture captured by the app during the first test. The passer-by in the photo has been blurred.

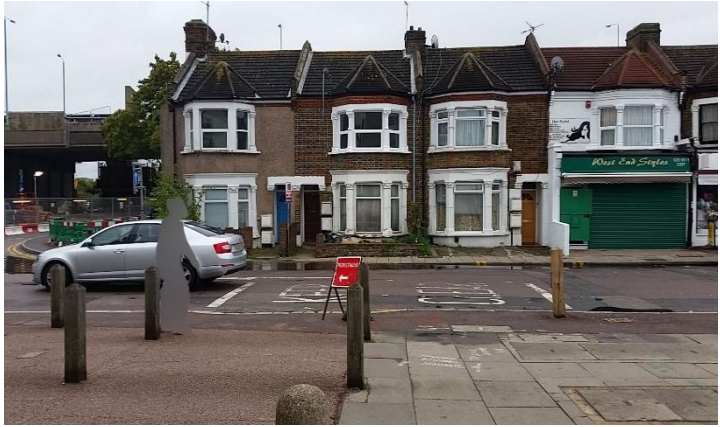

*Figure 5.1: A picture taken during first test*

The subsequent three pictures were captured during the second test:

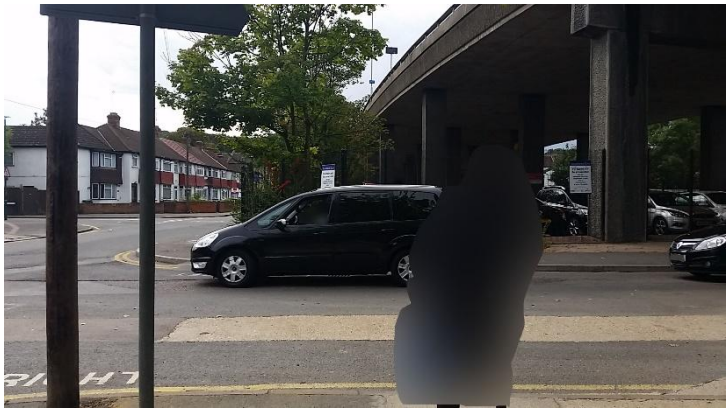

*Figure 5.2: Picture taken at 5miles/hr speed*

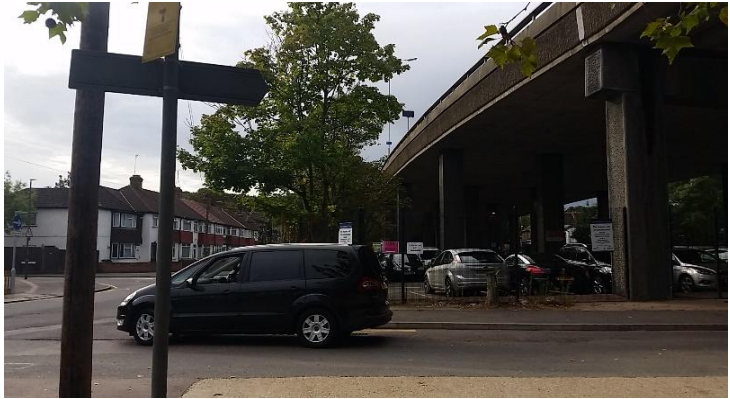

*Figure 5.3: Picture taken at 10miles/hr speed*

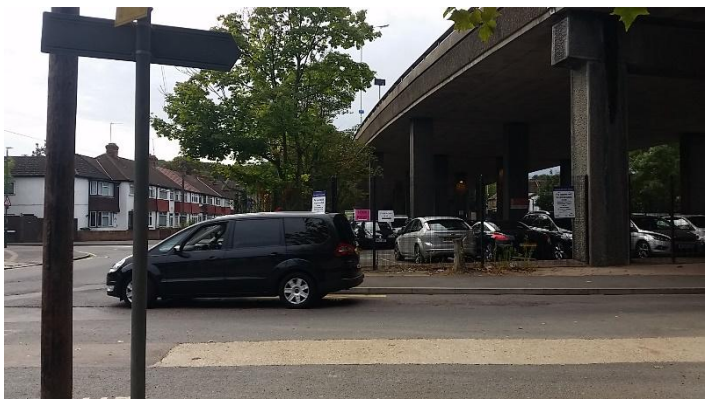

*Figure 5.4: Picture taken at 20miles/hr speed*

The table below summarises the reading of the app compared with the ground speed.

*Table 5.1: Ground speed versus app reading*

| Ground speed (mi/h) | App detected reading (0.25 x px/s) | detected reading (px/s) |
|---|---|---|
| 5 | 173.0 | 692 |
| 10 | 197.5 | 790 |
| 20 | 230.0 | 920 |

The dimensions used during the second test is shown below, the measurements were taken using blackspur® 10m tape measure:

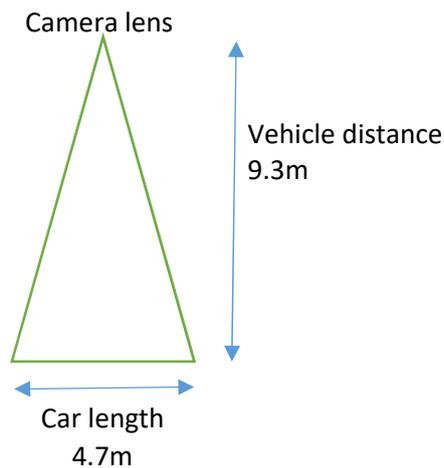

*Figure 5.5: Test distance dimensions*

The pixel length of the car was measured on Photoshop to be 704.58. Therefore the conversion coefficient at vehicle distance of 9.3m is 704.58/4.7 = 149.91px/m i.e. 1m equals 149.91pixels at a vehicle distance of 9.3m on a 1920x1080 frame picture. Applying the conversion coefficient to Table 5.1 gives the computed speeds as shown in Table 5.2.

*Table 5.2: Computed speed summary*

| Ground speed (mi/h) | Ground speed (km/h) | App detected reading (0.25 x px/s) | Detected reading (px/s) | Detected reading (m/s) | Detected reading (km/hr) |
|---|---|---|---|---|---|
| 5 | 8.0 | 173.0 | 692 | 4.6 | 16.5 |
| 10 | 16.1 | 197.5 | 790 | 5.3 | 19.1 |
| 20 | 32.2 | 230.0 | 920 | 6.1 | 30.0 |

From Table 5.2 it can be deduced that the arbitrary coefficient referred to in Section 4.2 which will better map the computed speeds in Figure 4.8 to ground truth speed is not 0.25 but rather 0.0033. That is using the reading with the least error, the arbitrary coefficient should be 30/920 = 0.0033

# 6    Conclusions and Recommendations

The Speed app project has succeeded in demonstrating that vehicular speeds could be measured from a smart phone camera. The detected speeds were proportional to the ground truth speeds. The discrepancy may be explained to arise from error in the reported ground truth speed, however the test vehicle used cruise control to maintain steady speeds.

To obtain the actual ground truth speed GPS speedometer may be preferable for accuracy as used in (Ginzburg et al., 2015), but it was not used in the test. The green detection rectangle fluctuated, recording displacement, even when a vehicle was stationary. This fluctuation explains why the reading for the lowest ground truth speed gives the greatest error or deviation from ground truth speed.

Furthermore, speed was computed using Euclidean distance applied to two dimensional axis x and y. Given the fluctuations a more reliable result could be obtained by considering only the changes in the x axis.

More stable detections have been observed when a larger volume of data is used to train the classifier. A case in point, is the default openCV face detector that ships with the SDK gives a stable detection and was trained with 3000 positive samples and 1500 negative samples in 19 stages. So stability in detections in the Speed app can be obtained by increasing the number of samples used in classifier training and improving the quality of the samples used in the training.

## 6.1    Knowledge Learnt

A lot has been learnt in the course of completing the Speed app project. The author had to personally acquire the knowledge to complete at least 80% of the project. The knowledge personally obtained to complete the project include: mastering the use of the DrawerLayout to design the visual interface of the MenuActivity so that menu options can be displayed on the drawer sliding list as shown in Figure 4.4. Making custom designs as seen in the layout buttons and the use of the Theme.Dialog theme to make the ResultActivity mimic an AlertDialog. These were all learnt in the course of the project. The codes to obtain geographic location were obtained from the official android site (Android Developers, 2016), however the codes were modularised into classes by the author for better readability and to make the app more object oriented. The manipulation of the Android Camera object, the saving of picture file to Android filesystem, the understanding and incorporation of the OpenCV 2.4.11 SDK for Android, the sending of large amount of data to a WCF service using HTTP POST method, the creation of the POST WCF service, the deployment of the WCF service on Microsoft Azure, the saving of data to the Azure Account storage (filesystem) from the WCF service, the linking of Microsoft SQL Management Server to Azure to view the live database on Azure and the accessing of the live database from the deployed WCF service were all personally studied in the development of the project.

## 6.2    Contribution to Knowledge

The classifier for the Speed app project has been trained using images from a standard database, i.e. the UIUC Image Database (Cogcomp.cs.illinois.edu, 2016). Therefore the reported fluctuations in detection of a stationary vehicle will alert other researchers to the challenges of using same image collection in their work. Furthermore, an observation has been that such fluctuations are minimised on a larger dataset.

The project has been successful in novel detection of vehicle speeds using computer vision, on a smartphone for a fixed vehicle distance. It has implicitly identified that for a full mobile detection experience which is independent of a vehicle distance, there is need for automatic re-computation of the conversion coefficient so that image distances are correctly mapped to vehicle distances.

With the increasing popularity of smartphones, an effective three dimensional (3D) mobile speed detector using computer vision can find wide application as a low cost alternative. This report is expected to generate further research and interest in perfecting the science of speed detection on mobile devices, by tackling the issues raised to further the development of the project.

The android code to make HTTP POST request to a C# WCF service and transferring JSON data where developed by the author. It has made available in this report to aid other programmers who need to implement a similar feature. Throughout the report, solutions to programming problems encountered in the project have been detailed.

### 6.3    Problems Encountered and Solution

Some of the problems that were encountered and their solution in the development of the project have already been discussed in the Implementation section. However, there were several more problems encountered during the project period. When the project topic was chosen the author had no idea of which API will eventually be used to implement the project and had to search on the Internet, before settling for OpenCV, other image recognition APIs were discovered. One of them was the SentiSight SDK for Android (Neurotechnology.com, 2016), the Sentisight SDK is an image recognition suite built on OpenCV. It was dersirable because it encapsulated the lower level detail that needed to be mastered to utilise the OpenCV library directly. However, it was dropped from the project for a number of reasons: It was a paid library and would have required a license to be used, while OpenCV is a free software, the sample projects in the SentiSight SDK had to be executed in Maven build while the author is conversant with the Gradle build for Android projects and the Sentisight library needed to be bundled together with the developed app, therefore resulting in a heavy deployment apk file (about 100MB) while OpenCV makes its library available to calling apps through a separately installed application which makes calling apps lighter and faster to build on IDE.

An epic solution was to develop the codes required to send POST data to the WCF web service, available solutions online used deprecated technologies which are no longer supported by Android and as stated in the Implementation section, after studying similar codes in C# and Ajax a solution was developed for Android and the WCF service through analogy.

Another problem encountered was that by default IIS server could only allow a maximum 4MB of data as request length, even after increasing the maxAllowedContentLength parameter of the requestLimits element it would still not accept more data. The solution was to also increase the maxReceivedMessageSize parameter value of the webHttpBinding element. Online solutions only made reference to maxReceivedMessageSize parameter under the basicHttpBinding element, while the Speed app does not use the basicHttpBinding but rather the webHttpBinding recommended for WCF. There was also the problem of under estimating the time needed to complete tasks but it was later realised that a distinction needed to be made between time estimated for a repeated task and for those that are novel to the author.

## 6.4   Changes to Project Plan

There have been some changes to the initial project plan. For example it was initially planned for the project web service to be deployed on the University's IIS server, but with the security challenges the University faced leading to the tightening of security policies it was a higher risk to continue the deployment on the University's server because accesses to the University's network was restrictive and would have required an active communication with the University's IT group to provide ordinary access. As an example, the author was unable to immediately run Android codes on the University's Android Studio IDE because Android Studio needed to download some plugins which was restricted by security setting. That occurrence had to be logged so that it could be resolved by the University's system admin group. Given the instability at the time it was preferable to rather deploy on Microsoft Azure.

## 6.5   Future Project Development

Much room for improvement exists for the Speed app. Currently, vehicle tracking has not been implemented in the Speed app. Future development on the project could focus on including tracking in the app. From the author's experience with Java tracking in OpenCV, a real time tracking will require calling not Java but the native C++ OpenCV library for Android to achieve real time performance. Due to the absence of tracking, the Speed app can only measure the speed of one vehicle at a time, but with tracking the speed of multiple vehicles can be tracked at a time.

With double camera lenses on the back of mobile devices, the lenses can be programmed to measure object distances through studio vision. Detected object distance value can be used to intermittently re-compute the pixel to actual distance ratio (conversion coefficient) on the Speed app which will ensure the accuracy of computed speed values for different vehicle distances.

Also there's need to improve the quality and increase the quantity of training data as has been discussed already at the beginning of this chapter.

# Appendices

## A.  Project Class Diagram

*Figure A.1: The Implementation Class Diagram*

## B.  Sequence Diagram for the Detect Vehicle Speed Use Case

*Figure B.1: The Detect Vehicle Speed Sequence Diagram*

## C. Sequence Diagram for the Obtain Location Information Use Case

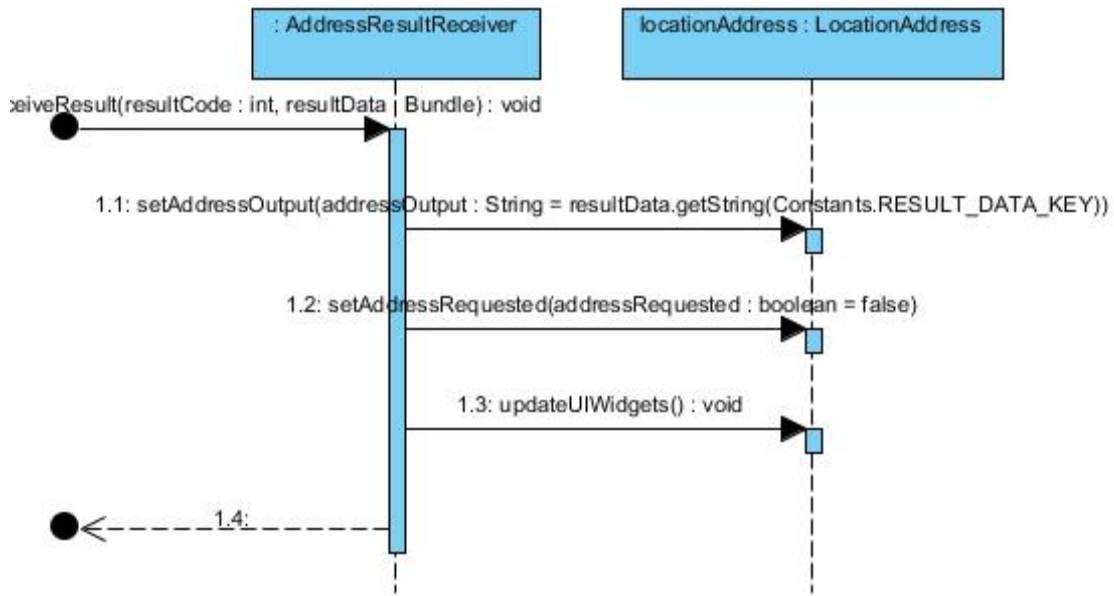

*Figure C.1: The Obtain Location Information Sequence Diagram*

## D. Sequence Diagram for the Capture Detection Time Use Case

Same as for Appendix B above.

# E. Sequence Diagram for the Take Vehicle Picture Use Case

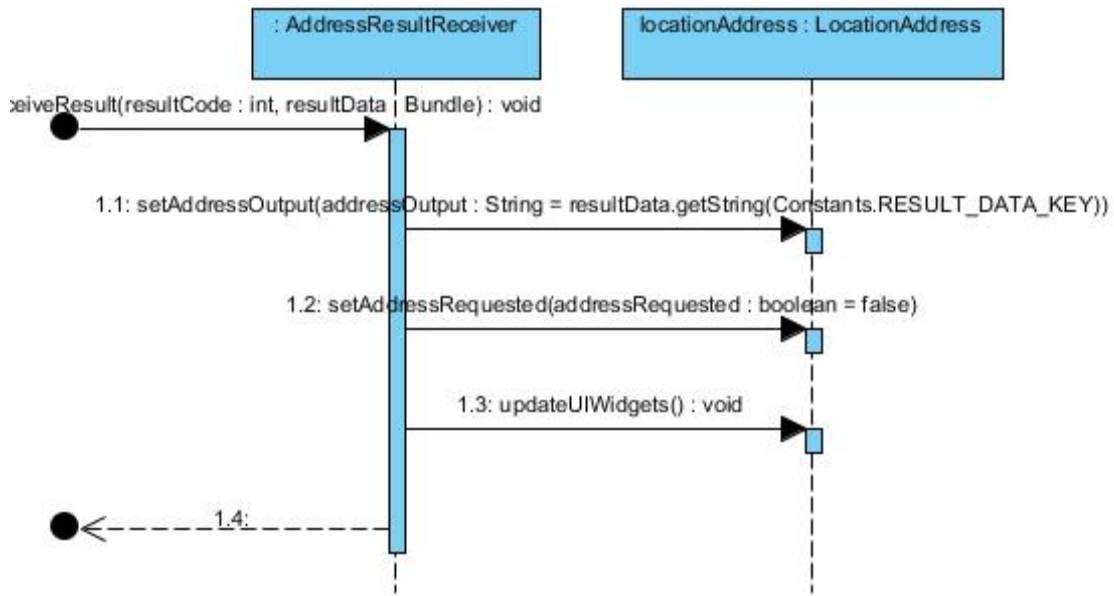

*Figure E.1: The Take Vehicle Picture Sequence Diagram*

## F. Sequence Diagram for the Save Generated Data Use Case

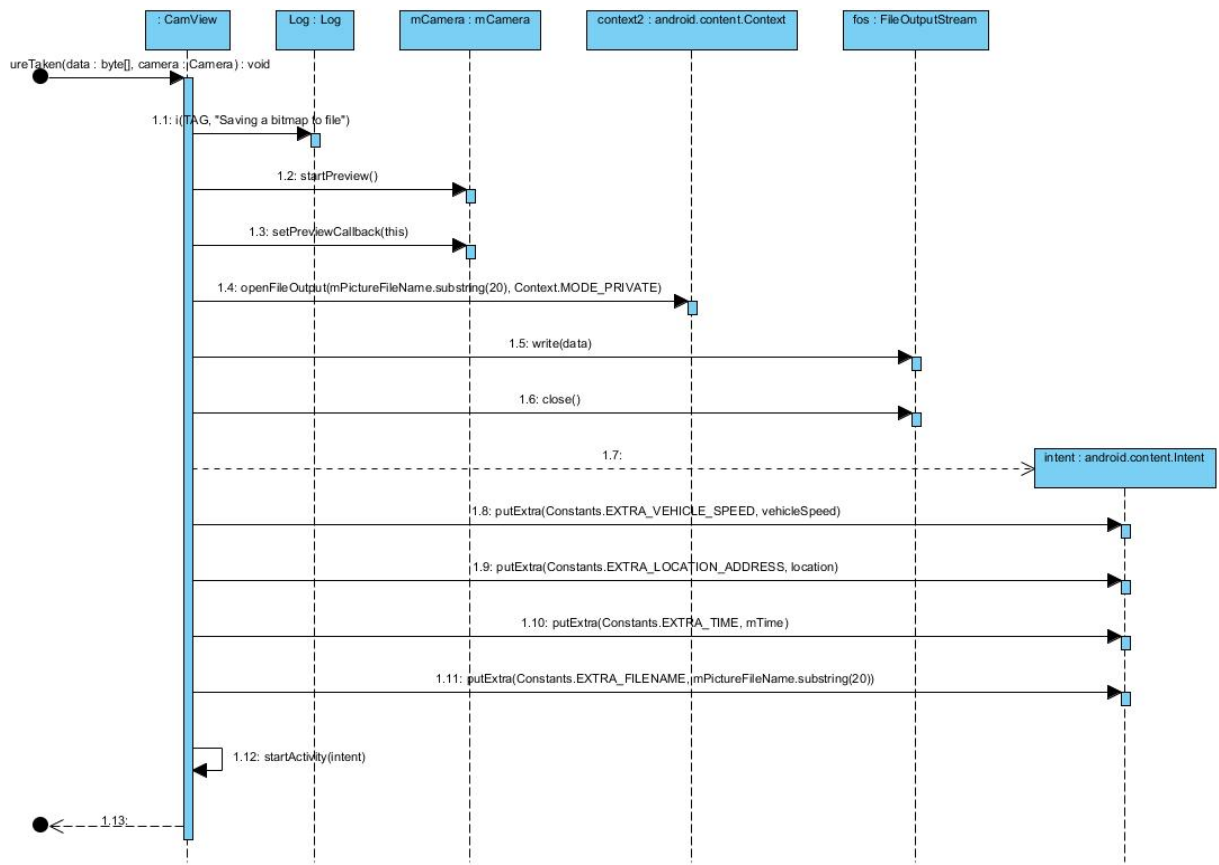

*Figure F.1: The Save Generated Data Sequence Diagram*

## G. Sequence Diagram for the View Saved Data Use Case

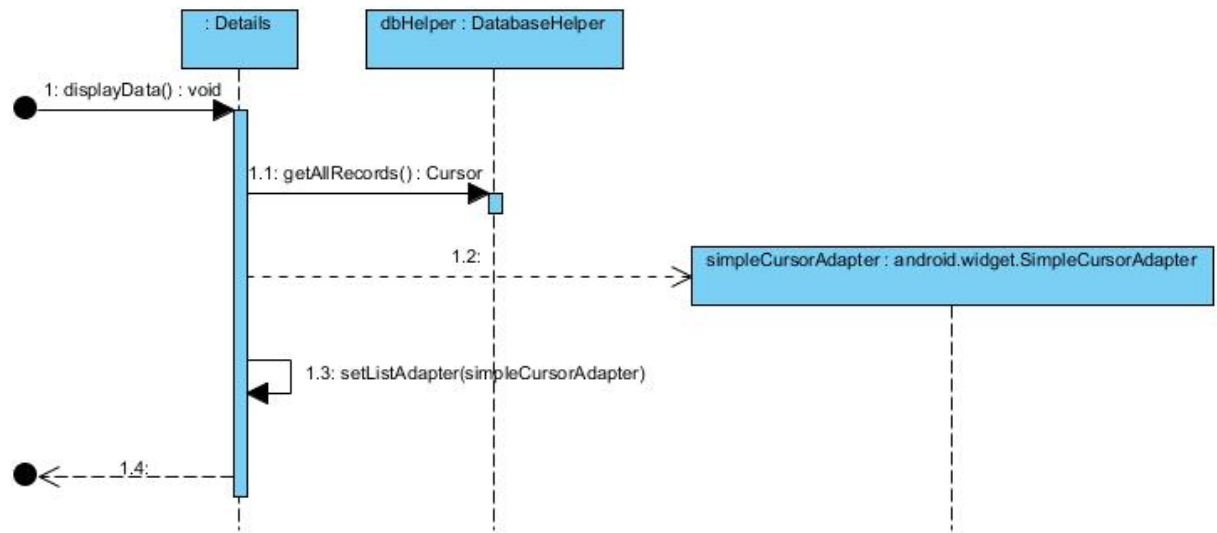

*Figure G.1: The View Saved Data Sequence Diagram*

## H. Sequence Diagram for the Search Saved Data by Date Field Use Case

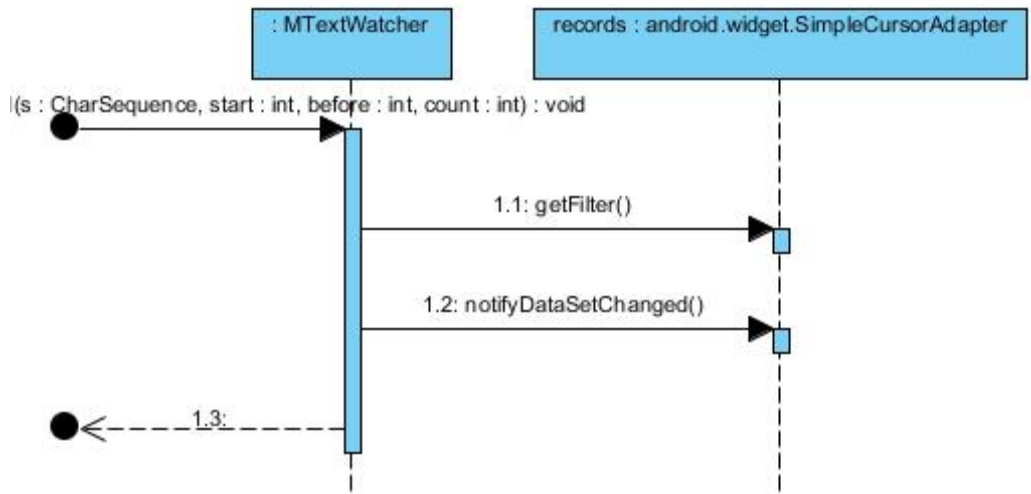

*Figure H.1: The Search Saved Data by Date Field Sequence Diagram*

# I. Sequence Diagram for the Delete Saved Data Use Case

*Figure I.1: The Delete Saved Data Sequence Diagram*

## J. Sequence Diagram for the Upload Saved Data to Web Server Use Case

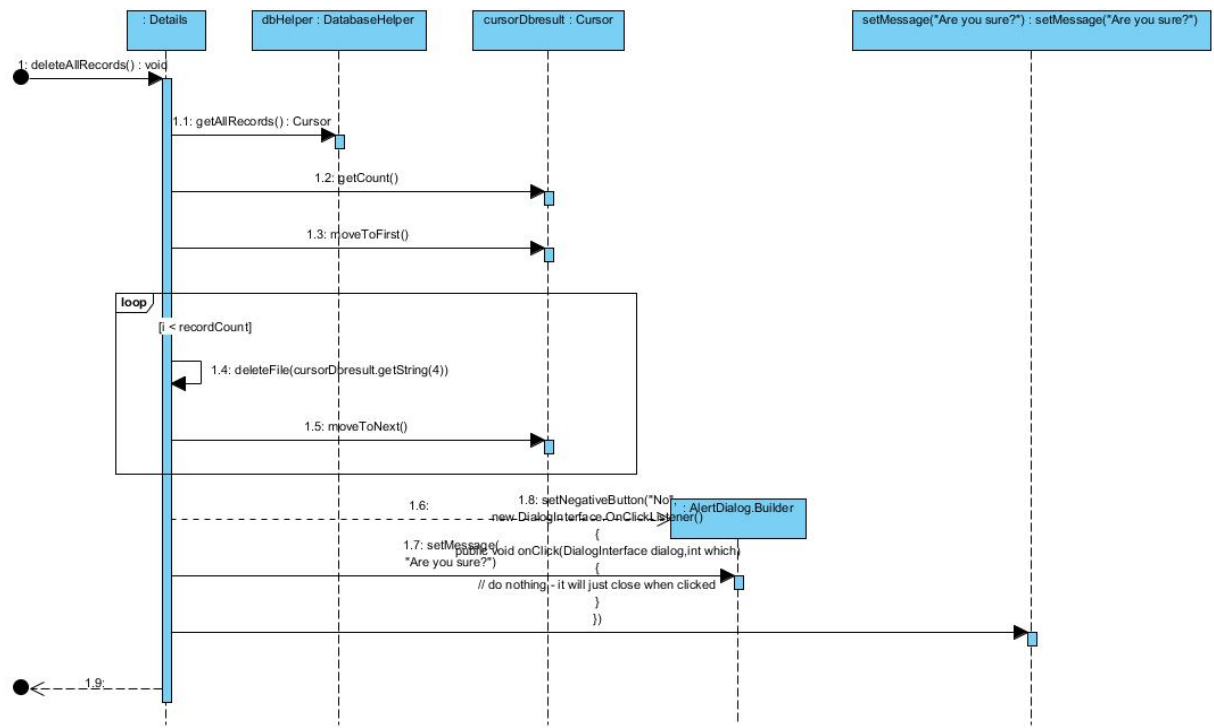

*Figure J.1: The Upload Saved Data to Web Server Sequence Diagram*